**Performance evaluation of large language model (LLM) for medication management tasks**

Kelli Henry, PharmD, MPA, BCCCP
kellirhenry@gmail.com
University of Colorado School of Medicine, Aurora, CO, USA

Steven Xu
shaochen.xu25@uga.edu
University of Georgia, Department of Computer Science, Athens, GA

Kaitlin Blotske, PharmD
Kaitlin.blotske@cuanschutz.edu
University of Colorado School of Medicine, Department of Biomedical Informatics, Aurora, CO, USA

Moriah Cargile, PharmD Candidate
MORIAH.CARGILE@cuanschutz.edu
University of Colorado Skaggs School of Pharmacy and Pharmaceutical Sciences, Aurora, CO, USA

Erin F. Barreto, PharmD, PhD, FCCM, FASN
Barreto.Erin@mayo.edu
Mayo Clinic, Rochester, MN, USA

Brian Murray, PharmD, BCCCP
BRIAN.2.MURRAY@CUANSCHUTZ.EDU
University of Colorado Skaggs School of Pharmacy and Pharmaceutical Sciences, Aurora, CO, USA

Susan Smith, PharmD, BCCCP, FCCM
Susan.Smith@uga.edu
University of Georgia College of Pharmacy, Athens, GA, USA

Seth R. Bauer, PharmD, FCCM, FCCP
bauers@ccf.org
Cleveland Clinic, Department of Pharmacy, Cleveland, OH, USA

Xingmeng Zhao, PhD
Xingmeng.zhao@ucdenver.edu
University of Colorado School of Medicine, Department of Biomedical Informatics, Aurora, CO, USA

Adeleine Tilley, PharmD Candidate
Adeleine.tilley@cuanschutz.edu
University of Colorado Skaggs School of Pharmacy and Pharmaceutical Sciences, Aurora, CO, USA

Yanjun Gao, PhD
Yanjun.Gao@cuanschutz.edu
University of Colorado School of Medicine, Department of Biomedical Informatics, Aurora, CO, USA

Tianming Liu, PhD

tianming.liu@gmail.com
University of Georgia, Department of Computer Science, Athens, GA, USA

Sunghwan Sohn, PhD
Sohn.Sunghwan@mayo.edu
Mayo Clinic, Department of Artificial Intelligence and Informatics, Rochester, MN, USA

Andrea Sikora, PharmD, MSCR, BCCCP, FCCM, FCCP
andrea.sikora@cuanschutz.edu
University of Colorado School of Medicine, Department of Biomedical Informatics, Aurora, CO, USA
sikora@uga.edu
University of Georgia College of Pharmacy, Department of Clinical and Administrative Pharmacy, Athens, GA, USA

**Conflicts of Interest:** The authors have no conflicts of interest.

**Funding:** Funding: Funding through Agency of Healthcare Research and Quality for Drs. Sikora, Smith, and Liu was provided through R21HS028485 and R01HS029009. Funding through National Institutes of General Medical Sciences for Dr. Bauer was provided through K08 GM147806.

**Acknowledgements:** Christopher Barry

**Abstract**

**Purpose:** Large language models (LLMs) have proven performance for certain diagnostic tasks, however limited studies have evaluated their consistency in recommending appropriate medication regimens for a given diagnosis. Medication management is a complex task that requires synthesis of drug formulation and complete order instructions for safe use. Here, the performance of GPT-4o, an LLM available with ChatGPT, was tested for three medication management tasks.

**Methods:** GPT-4o performance was tested using three medication tasks: identifying available formulations for a given generic drug name, identifying drug-drug interactions (DDI) for a given medication regimen, and preparing a medication order for a given generic drug name. For each experiment, the model's raw text response was captured exactly as returned and evaluated using clinician evaluation in addition to standard LLM metrics, including Term Frequency-Inverse Document Frequency (TF-IDF) vectors, normalized Levenshtein similarity, and Recall-Oriented Understudy for Gisting Evaluation (ROUGE-1/ROUGE-L F1) between each response and its reference string.

**Results:** For the first task of drug-formulation matching, GPT-4o had 49% accuracy for generic medications being matched to all available formulations, with an average of 1.23 omissions per medication and 1.14 hallucinations per medication. For the second task of drug-drug interaction identification, the accuracy was 54.7% for identifying the DDI pair. For the third task, GPT-4o generated order sentences containing no medication or abbreviation errors in 65.8% of the cases.

**Conclusions:** Model performance for basic medication tasks was consistently poor. This evaluation highlights the need for domain-specific training through clinician-annotated datasets and a comprehensive evaluation framework for benchmarking performance.

## Introduction

The most recent large language models (LLMs) have demonstrated notable proficiency for a variety of medical tasks including passing medical licensing exams and selecting correct medical diagnoses.[1, 2] However, limited evaluations have analyzed LLM performance for the safety and efficacy in medication-related tasks. Generating safe and effective medication regimens requires specialized domain knowledge and human-like reasoning to synthesize drug product information with patient specific factors. This complex process is generally termed comprehensive medication management (CMM).[3, 4] While information technology (IT) solutions have had dramatic impacts in making medication use safer (e.g., computerized medication order entry, barcoding to match drug to patient), few artificial intelligence (AI) based tools have been developed with CMM in mind.

In early analyses, LLMs have shown promising results for medication tasks, yet this promise comes with notable performance gaps that require improvement prior to safe clinical deployment. In order to safely take medications, there are essential pieces of information necessary, including the exact dosing form of a medication (e.g., aspirin 81mg tablet) in addition to the dose, route, and frequency (e.g., one tablet by mouth daily). Given the variety of medications available, each piece of information is dependent upon the other and requires both knowledge and reasoning for safe use. Our team has conducted several exploratory analyses in this space evaluating performance on simple, well-structured medication questions in the form of Doctor of Pharmacy multiple choice exam questions, identification of clinically relevant drug-drug interactions, and performance on a small series of complex patient cases.[4-6] In each case, while LLMs showed marked insight into the complexities of medications, there were concerning and even life-threatening recommendations and outputs. As such, the need to systematically characterize LLM performance prior to clinical deployment is essential.

The purpose of this analysis was to determine the current performance of an established high-performing LLM, GPT-4o (an updated version of GPT-4 available to ChatGPT), on three medication tasks that are considered essential to safe medication use: identification of all available drug formulations when provided the generic drug name, provision of a full medication order (dose, unit, route, and frequency) when provided the generic drug name, and identification of drug-drug interactions within a list of 3-6 medications.

## Methods

Given the inherent complexity of CMM, this performance evaluation was focused on three baseline medication tasks that were deemed essential by the clinician panel for safe and efficacious deployment in a healthcare setting: (1) drug formulation identification, (2) medication order or sig generation, and (3) drug-drug interaction identification. While not fully representative of CMM, these tasks are absolutely essential for patient safety any time medications are used in the healthcare setting. A high-performing LLM (GPT-4o) was selected: given the fast pace of LLM model advancement, this LLM was chosen for its established performance, with the intent to establish that these specific medication tasks are reasonable performance indicators for LLM-based medication tasks. This project was reviewed and approved by the University of Colorado Institutional Review Board (COMIRB #25-1631). All methods were performed in accordance with the ethical standards of the Helsinki Declaration of 1975. This is summarized in **Figure 1**.

All experiments involved a structured prompting process to elicit an exact response from the LLM. All medications were queried in the summer of 2025 using OpenAI Application Programming Interface (API).[7] GPT-4o was used with the default settings of Temperature=1.0 and Top P=1.0. Temperature and Top P allow for variation in creativity (also known as randomness) with higher values allowing for more diverse and less predictable text.[8] A temperature and Top P of 1 allows the LLM to use all words in its vocabulary.[8] Each prompt was queried one at a time so that contextual carry-over could not influence successive generations. Results were analyzed using Microsoft Excel (version 2025).

***Task 1: Drug Formulation Identification***

*Study Design*
The primary objective was to evaluate the ability of GPT-4o to match the generic drug name to all available drug formulations. The primary outcome was the percentage of medications that were returned with a complete list of available formulations. A list of available formulations was generated by referencing LexiDrug™ for each medication.[9] The list of medications (n=290) was generated from a list of most prescribed medications in the United States.[10] The complete prompt given to GPT-4o was:

"What are all of the formulations (including the numerical value and the unit) available in the United States for the drug '{drug_name}'? ONLY return the formulation and associated unit in a format as such: 200 mg tablet; 400 mg capsule; 800 mg extended release tablet; 50 mg/mL injectable solution; 50 mg/mL intramuscular suspension; 2% cream. DO NOT return any other text!"

*Data Analysis*
The primary outcome was the overall accuracy rate (correct formulations minus hallucinated formulations divided by total available formulations). Descriptive statistics were performed to calculate total errors, total hallucinated formulations, and missing formulations per medication. Additionally, an exploratory analysis using Mann Whitney U was conducted to compare LLM accuracy in formulation identification for medications with many formulations (5-10 or >10) compared to medications with fewer formulations (<5). Formulations that were identified but do not exist were termed hallucinations. Total formulations were defined as all available formulations (4% cream and 5% cream count as separate formulations). Unique types of formulations were defined as separate formulations by route and type (4% and 5% cream would count as 1 formulation "cream" which would be different than an ointment or topical liquid).

***Task 2: Medication Order or Sig Generation***

*Study Design*
The primary objective was to evaluate the ability of GPT-4o to match a drug name to an appropriate medication order including formulation, dose, route, and frequency as both an abbreviated "prescription sig" and an expanded "Complete Instructions." In order to safely take medications, there are essential pieces of information necessary, including the exact dosing form of a medication (e.g., aspirin 81mg tablet) in addition to the dose, route, and frequency (e.g., one tablet by mouth daily). Given the variety of medications available, each piece of information is dependent upon the other and requires both knowledge and reasoning for safe use. The “sig” included all aspects of a complete medication order but uses abbreviations (e.g. pregabalin 75mg PO BID); whereas the “complete instructions” does not include abbreviations (e.g. pregabalin 75mg take one capsule by mouth twice daily). A few-shot prompt was crafted by embedding five randomly selected examples (each with name, correct Sig, and correct complete instructions) at the top of the conversation, followed by explicit instructions that every output must follow the order *drug → dose → amount → route → frequency*. The primary outcome was the percentage of instructions that had drug errors (dose, formulation, route, frequency). Secondary outcomes included percentage of instructions with a lack of standardization of abbreviations or inappropriate abbreviations used. A total of 38 drugs were tested. The complete prompt given to GPT-4o was:

“You are a medical assistant tasked with providing the correct prescription instruction when given a drug name.
You must provide the output in two formats:
1. A simplified instruction.
2. A clarified version of the instruction.
Use the following training examples to understand the format:
Drug Name: {row['Generic Drug Name']}
Simplified Version: {row['Prescription Bank Sig'].strip()}
Clarified Version: {row['Clarification'].strip()}
Now, using the same format and strictly following the structure of drug name followed by dose, intake amount, intake method, and frequency as seen above, provide the simplified and clarified instructions for the following drug.
Drug Name: {drug_name}”

The prompt was part of a dynamic code that randomly selected 5 samples from the ground truth test bank (**Supplementary Appendix**) to give as an example. For example,
Drug Name: amlodipine
Simplified Version: Amlodipine 2.5mg PO once daily
Clarified Version: Amlodipine 2.5mg Take one tablet by mouth once daily

The predicted Prescription Sig (simplified version) and Complete Instructions (clarified version) were scored independently by two board certified pharmacists who reviewed the predicted sig and instructions for any discrepancies and/or omission of information, including the appropriate use of abbreviations. Abbreviations not aligning with guidance from the Institute of Safe Medication Practices (ISMP) on best practices for abbreviations were considered inappropriate abbreviations.[11] Any discrepancies between the two primary graders were resolved with a third adjudicator, also a board-certified pharmacist. The primary outcome was percentage of generated sig with no errors. Secondary outcomes included percentage of generated sig with a drug error or an abbreviation error. Drug errors and abbreviation errors were counted as categorical variables.

***Task 3: Drug-Drug Interactions***

The primary objective was to evaluate the ability of GPT-4o to correctly identify if a list of 3-6 medications contained no drug interactions or one drug interaction between two medications and list the medications involved in that interaction. The primary outcome was the percentage of matching responses to the "ground truth".[9]

*LLM Testing*
A total of 83 medication lists containing 3-6 medications each along with complete instructions for the medications (drug, dose, formulation, route, and frequency) were tested for drug interactions. A "ground truth" answer key was developed using the LexiDrug™ drug-drug interaction checker.[9] GPT-4o was prompted with the following:

"Is there a drug-drug interaction present in this query of medications? If so, what are the two interacting drugs, respond with 'DrugA and DrugB', otherwise, respond with 'No interaction'. Your response must be only the drug pair or 'No interaction'! Do not include anything else!"

Each medication list was queried five times using GPT internal knowledge (gpt-4o model) and five times where GPT was permitted to use a single web search to aid in its response (gpt-4o-search-preview). This was done to prevent the tendency of LLMs for sycophancy, with a tendency to say "yes" if asked a question.[12]

Descriptive statistics for both the internal knowledge and web-assisted responses were utilized, including consistency of the responses, identification of medications not on the medication list, and overall accuracy. Accuracy was quantified by computing cosine similarity on Term Frequency–Inverse Document Frequency (TF-IDF) vectors, normalized Levenshtein similarity, and Recall-Oriented Understudy for Gisting Evaluation (ROUGE-1/ROUGE-L F1) between each response and its reference string.[13-15] These scores were calculated per drug and averaged to yield a global measure for each model variant. The ROUGE (Recall-Oriented Understudy for Gisting Evaluation) score was calculated by comparing the model-generated output with the given example cases.[13] The ROUGE score is an evaluation metric widely used in the field of natural language processing (NLP), especially in text summarization and machine translation.[13] It is mainly used to evaluate the quality of automatically generated summaries or translations.[13] The core idea of ROUGE is to evaluate candidate abstracts by comparing them to a set of reference abstracts (usually human-written) by measuring overlap of n-grams (or word sequences that appear in both summaries) with recall defined as number of n-grams appearing in both summaries divided by total number of n-grams in the reference summary.[13] Values range from 0-1, with higher values indicating better summary quality.[13] Descriptive statistics were performed, including Chi Square and Mann Whitney U for categorical and continuous variables as appropriate.

$$ROUGE-N = \frac{\sum_{gram_n \in Reference\ Summaries} Count_{match}(gram_n)}{\sum_{gram_n \in Reference\ Summaries} Count(gram_n)}$$

**Data Availability**
The data underlying this article will be shared on reasonable request to the corresponding author.

## Results

*Drug Formulation Identification*
A total of 290 medications were tested, with GPT-4o correctly identifying all available formulations of 138 (47.6%) medications. The overall accuracy rate (correct formulations minus hallucinated formulations divided by total formulations) was 65.6%. Per medication, there was a median (interquartile range) of 1 (0-3) errors, reflective of 0 (0-1) missing formulations, and 0 (0-1) hallucinations (i.e. formulations not known to be available). In exploratory analysis, there was evidence for an increase in error rate with increasing number of available formulations as well as unique types of formulations (**Supplementary Appendix**). Full results are detailed in **Tables 1 and 2**.

*Full Medication Order Generation*
A total of 38 medications were tested with GPT-4o: 6 medication orders generated by the LLM contained at least 1 drug error (15.7%) while eight medication orders (21.1%) contained at least one lack of standardization of abbreviations or use of an inappropriate abbreviation. Only 25 generated prescription sigs contained no medication or abbreviation errors (65.8%). A description of the generated sig and errors associated is included in **Table 3.** The LLM repeatedly used abbreviations that are not recommended by ISMP, including “OU” instead of “both eyes”, “SC” instead of “SUBQ”, “1 T” instead of “1 tablet”, and “Pl” instead of “place.”[11] Additionally, the LLM hallucinated an abbreviation of “QIDAC” which would mean “four times daily with meals;” however, this is not a standard order format.

*Drug-Drug Interactions*
GPT-4o had a ROGUE-1 FI average of 55.2% using the internal knowledge test and a ROGUE-1 F1 average of 66.3% (**Table 4**) when allowed a single web search prior to its response. Variances with each test ranged from 1.78 to 7.81. The mean percentage of correct answers was 54.7% (±43.4) in the internal knowledge query compared to 69.2% (±39.8) in the web-assisted query (p=0.013) (**Table 5**). GPT-4o performed better with the web-assisted query for Category C and Category D interactions but performed worse with drug combinations that included no drug interaction (**Table 5**).

**Discussion**

In the first systematic evaluation of LLM performance for medication tasks representative of CMM and the medication use process, a generally high-performing LLM (GPT-4o) showed marked need for performance improvement before it can safely be used for such tasks. GPT4-o was unable to consistently match drug name with available products and had notable hallucinations. Moreover, performance was low for DDI identification, which is particularly important given that non-AI based information technology tools are highly reliable interfaces for clinicians to identify DDIs. Additionally, errors were present in LLM-generated medication order sentences, including some dosing errors (e.g., underdosing a medication frequency) as well as formulation errors (e.g., dosing frequency is incorrect based on type of product selected [i.e. extended-release, immediate-release, delayed-release]). These errors have significant potential for harm in real-world settings, and importantly are not issues with standardized electronic health record formats because formulations and DDIs are hard-coded into those systems.

These tasks represent a small portion of the required components for clinical evaluation of medications. Medication decisions integrate a complex knowledge base including synergistic mechanisms of drug action and effects of acute illness on pharmacokinetic parameters and pharmacodynamic response.[16] The alphanumeric text involved with medication orders (e.g., cefepime 2g intravenously every 8 hours) requires highly specialized domain knowledge to interpret, particularly in the context of an individual patient (e.g., is this dose appropriate for their present kidney function?) and other medications they are receiving (e.g., is this new medication safe to give with other antibiotics they are taking?). An algorithm must further discern whether drug dose affects adverse drug event risk (e.g., cefepime neurotoxicity is dose-dependent, whereas a cefepime allergic reaction is not). The AI algorithm must recognize and apply a knowledge of other patient specific information like the impact of dialysis (e.g., cefepime dose reductions in renal failure are appropriate, but when continuous dialysis is initiated for that renal failure, the dose should actually be higher than the dose typically used in patients with renal failure). These issues have precluded appropriate AI-based medication-decision making capacity because incorrect medication decision-making has life-threatening implications.[17, 18] LLMs do not appear to have appropriate domain-specific training to reliably interpret and recommend medications, and safe deployment will likely require essential break down of the medication use process into a testable series of steps.[17, 18]

In evaluating drug-drug interactions, GPT-4o was inconsistent in its answers, often giving multiple different answers when queried several times. While a web search improved performance, similarities between the LLM response and the ground truth were not as consistent as desired. Additionally, the web search feature actually worsened identification of scenarios with no DDIs. This may serve as an example of LLMs having a bias towards sycophancy, such that asking if a DDI is present has a likelihood of bringing back a positive answer.[12] Given that DDI information has been integrated effectively into present-day electronic health record and other health IT tools, this represents an important gap for future evaluations to fill. Notably, a retrieval-augmented generation (RAG) strategy has been shown to improve LLM performance by integrating external data into the LLM system.[19-21] This analysis used a form of this strategy by allowing the LLM to perform a single web search prior to completing the request. However, the LLM still failed to correctly identify the drug-drug interaction. This may indicate that while web-based searches can be helpful for other LLM queries, CMM is a complex topic that requires specific and intentional training for LLMs to be capable of consistently generating appropriate responses.[22-24]

Considering that the majority of LLMs are trained on a widely available corpus (e.g., the Internet), there are likely additional challenges in specialties including highly technical language or uncommon situations, which is often the case in medicine overall and specifically the field of pharmacy.[25, 26] Future RAG approaches based on a trust drug information source would likely be beneficial in improving the accuracy of LLM responses.

There is a clear need for clinically relevant, transparent benchmarks to judge the safety and efficacy of LLM-based medication recommendations. While clear standards exist for clinicians (e.g., board certification, graduate degrees) and predictive models (e.g., positive and negative predictive value), LLMs have not been subjected to the clinical rigor necessary for this technology to be safely used at the bedside. It can be quite difficult to calculate or even find documented error rates, hallucination rates, and accuracy, making it difficult to understand how much a user can trust the output. A recent viewpoint by the United States Food & Drug Administration explicitly stated the need for "external stakeholders to ramp up assessment and quality management of AI."[27] Given the resource intensive nature of clinician annotation, an automated means of evaluating the accuracy of medication recommendations is ultimately necessary for LLMs to reach the safety thresholds necessary for bedside use, and a standardized clinician-annotated benchmark dataset may improve this development process.

This series of medication tasks represents one of the first attempts to create standardized means of evaluating LLM performance, particularly as it relates to safety. While this evaluation was intended as a proof-of-concept analysis to demonstrate the need for a systematic evaluation using established performance indicators, there are limitations of this analysis that preclude immediate uptake. Further analyses are absolutely essential for development and include testing multiple LLMs, employing various prompting and reasoning strategies, and use of multiple iterations of LLM output for reliable evaluation of self-consistency and overall performance. These types of rigorous analysis are ultimately necessary prior to the safe deployment of LLMs for healthcare tasks in a patient-facing scenario. Moreover, larger, clinician annotated datasets are an important next step. This analysis serves as important evidence to support these larger, more resource intensive methods.

## Conclusions

These three medication tasks identified lack of consistency and accuracy when creating medication order sentences and identifying formulations, a tendency to omit and hallucinate available formulations, and a tendency to identify drug-drug interactions that do not exist. Further optimization of LLM performance using robust benchmarking is necessary prior to consistent use of LLMs at the bedside for medication tasks.

**Figure 1. Methodological Process**

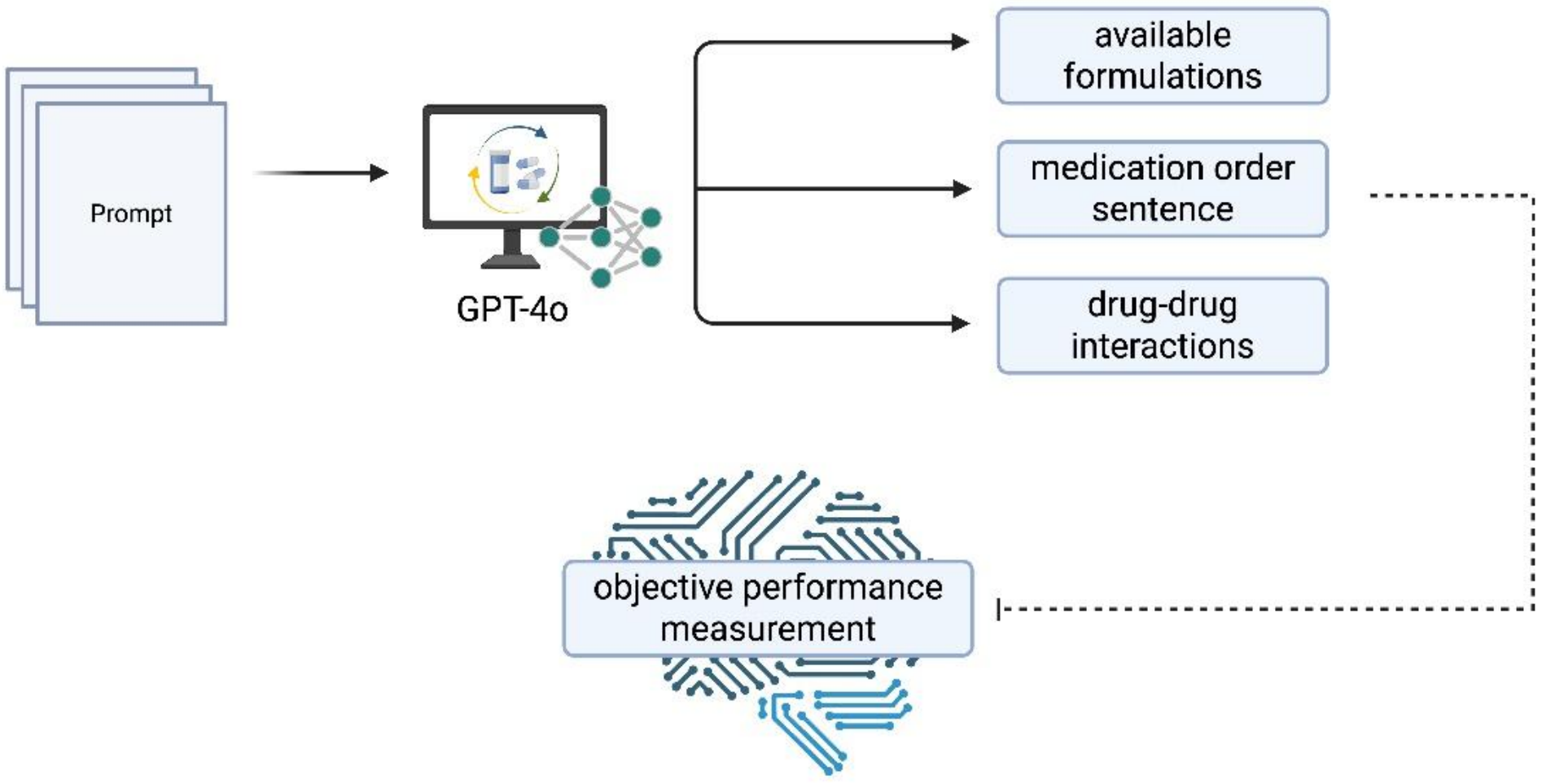

**Table 1. Drug Formulation Test Bank**

| Feature | N=290 |
|---|---|
| Total formulations per medication, mean ± SD | 7.1 ± 6 |
| Unique types of formulations per medication, mean ± SD | 2.7 ± 1.9 |
| Total formulations, n | 2088 |
| Formulation Types, n (%) | |
| Oral | 1486 (71.2) |
| Tablet | 636 (30.5) |
| Extended-release capsule or tablet | 298 (14.3) |
| Capsule | 189 (9.1) |
| Oral solution, elixir, or concentrate | 113 (5.4) |
| Orally disintegrating, effervescent, or sublingual | 61 (2.9) |
| Oral suspension | 55 (2.6) |
| Delayed-release tablet or capsule | 46 (2.2) |
| Chewable capsule or tablet | 28 (1.3) |
| Oral packet or powder | 18 (<1) |
| Oral granules or pellets | 11 (<1) |
| Oral film | 6 (<1) |
| Sublingual liquid | 5 (<1) |
| Lozenge | 4 (<1) |
| Extended-release chewable tablet | 3 (<1) |
| Extended-release orally disintegrating tablet | 3 (<1) |
| Sustained-release tablet or capsule | 3 (<1) |

| Film-coated tablet | 2 (<1) |
|---:|---|
| Extended-release oral suspension | 2 (<1) |
| Oral syrup | 2 (<1) |
| Buccal tablets or capsules | 1 (<1) |
| Intravenous | 190 (9.1) |
| Injectable solution or suspension | 170 (8.1) |
| Injectable syringe | 14 (<1) |
| Injectable powder | 6 (<1) |
| Intramuscular | 58 (2.8) |
| Intramuscular suspension or solution | 33 (1.6) |
| Intramuscular oil | 12 (<1) |
| Extended-release intramuscular suspension | 11 (<1) |
| Intramuscular syringe or autoinjector | 2 (<1) |
| Subcutaneous | 38 (1.8) |
| Subcutaneous solution or suspension | 36 (1.2) |
| Subcutaneous pellet or implant | 1 (<1) |
| Extended release subcutaneous solution | 1 (<1) |
| Inhalation | 56 (2.7) |
| Inhalation powder, capsule, or cartridge | 21 (1) |
| Inhalation solution, suspension, or concentration | 15 (<1) |
| Inhalation aerosol | 14 (<1) |
| Metered dose inhaler | 6 (<1) |
| Intrathecal solution | 4 (<1) |
| Intraarticular suspension | 2 (<1) |
| Nasal | 23 (1.1) |
| Nasal spray, solution, or powder | 21 (1) |
| Implant | 1 (<1) |
| Ointment | 1 (<1) |
| Topical | 159 (7.6) |
| Cream | 36 (1.7) |
| Gel | 30 (1.4) |
| Ointment | 22 (1.1) |
| Transdermal patch | 21 (1) |
| Topical solution or external liquid | 20 (1) |
| Lotion | 15 (<1) |
| Foam | 5 (<1) |
| Topical spray or aerosol | 4 (<1) |
| Shampoo | 2 (<1) |
| Topical powder | 1 (<1) |
| Dental paste | 1 (<1) |
| Pad | 1 (<1) |
| External swab | 1 (<1) |
| Ophthalmic | 44 (2.2) |
| Solution | 25 (1.2) |
| Suspension or emulsion | 13 (<1) |
| Gel | 3 (<1) |
| Intracameral or ophthalmic implant | 2 (<1) |
| Ointment | 1 (<1) |
| Otic Solution | 2 (<1) |
| Rectal | 15 (<1) |
| Suppository | 9 (<1) |
| Gel | 4 (<1) |
| Foam | 1 (<1) |
| Ointment | 1 (<1) |
| Vaginal | 9 (<1) |
| Gel | 5 (<1) |
| Suppository | 2 (<1) |
| Cream | 1 (<1) |
| Ring | 1 (<1) |

Total formulations = all available formulations (4% cream and 5% cream count as separate formulations). Unique types of formulations = separate formulations by route and type (4% and 5% cream would count as 1 formulation "cream" which would be different than an ointment or topical liquid). Types of formulations are detailed in Table 1.

**Table 2**. Drug Formulation Identification Results

| | N=290 |
|---|---|
| Total errors per medication | 1 (0-3) |
| Missing formulations | 0 (0-1) |
| Hallucinations | 0 (0-1) |
| Medications with completely correct formulations | 138 (47.6) |
| Medications with 1 total error | 47 (16.2) |
| Medications with >1 total errors | 105 (36.2) |
| Total missing formulations, n | 358 |
| Total hallucinated formulations, n | 330 |
| Overall accuracy | 1369 (65.6) |

All values n (%) or median (interquartile range) unless otherwise stated.

**Table 3.** Full Medication Order Generation Drug Errors

| Medication Prompt (Provided to LLM) | Prescription Sig (Generated by LLM) | Error Description |
|---|---|---|
| Fluticasone/ Salmeterol | Fluticasone/ Salmeterol 100 mcg/50 mcg Inh 1 puff BID | The dosage form selected is a dry powder inhaler so the dose should say 1 inhalation not 1 puff as this medication cannot be puffed. |
| Bupropion | Bupropion 150 mg PO BID | Bupropion immediate release was selected which is not available as a 150 mg tablet and should have been dosed 3-4 times per day. This would be appropriate if the sustained release dosage form was selected which is available as a 150 mg tablet and is dosed q12h. |
| Diltiazem | Diltiazem 60 mg PO BID | Diltiazem immediate release was selected which should have been dosed 4 times per day. This would be appropriate if the 12-hour extended-release formulation was selected. |
| Diphenoxylate/ Atropine | Diphenoxylate/Atropine 5 mg/0.05 mg PO 1 T QID PRN | There are two errors identified. 1. These tablets come as 2.5/0.025 so this should be 2 tablets to make this dose. 2. It should say PRN diarrhea instead of just PRN. |
| Hydralazine | Hydralazine 25 mg PO BID | Hydralazine is dosed at least 3-4 times per day, so this is an underdosing of this medication. |
| Latanoprost | Latanoprost 1 drop in affected eye(s) QHS | No strength listed- should be 0.005%. |

BID: twice daily, LLM: large language model, mcg: microgram, mg: milligram, PO: by mouth, PRN: as needed, q12h: every twelve hours, QHS: every night at bedtime, QID: four times daily

**Table 4.** Drug-Drug Interaction Similarity Evaluation

| | GPT-4o Internal Knowledge Query*[13-15] | GPT-4o Web-assisted Query**[13-15] |
|---|---|---|

| Attempts | Cosine Similarity | Levenshtein | ROGUE-1 F1 | ROGUE-L F1 | Cosine Similarity | Levenshtein | ROGUE-1 F1 | ROGUE-L F1 |
|---|---|---|---|---|---|---|---|---|
| 1 | 47.5% | 54.8% | 57% | 50.9% | 65.1% | 53% | 64% | 51.1% |
| 2 | 43.5% | 50.6% | 53% | 47.0% | 69.2% | 57.8% | 68.3% | 57.1% |
| 3 | 46.2% | 51.5% | 55.5% | 48.2% | 67.6% | 54.4% | 67.5% | 53.7% |
| 4 | 42.6% | 47.6% | 51.7% | 43.4% | 67.7% | 57% | 67.9% | 56% |
| 5 | 49.4% | 53.6% | 58.8% | 50.9% | 67.5% | 53.6% | 63.9% | 51.3% |
| Overall Average | 45.8% | 51.6% | 55.2% | 48.1% | 67.4% | 55.1% | 66.3% | 53.8% |
| Variance | 6.3 | 6.2 | 6.8 | 7.8 | 1.8 | 3.6 | 3.9 | 5.9 |

*used gpt-4o model

**used gpt-4o-search-preview

**Table 5.** Drug-Drug Interaction Results

| | Internal Knowledge Query* (n=83) | Web-assisted Query** (n=83) | P value |
|---|---|---|---|
| Percentage of correct answers, mean (±SD) | 54.7 (43.4) | 69.2 (39.8) | 0.01 |
| Percentage of correct answers | 60 (0-100) | 100 (40-100) | 0.04 |
| Queries with >1 unique answers | 33 (39.8) | 28 (33.7) | 0.42 |
| Queries with >2 unique answers | 10 (12) | 5 (6) | 0.28 |
| Queries that were completely correct | 33 (39.8) | 45 (54.2) | 0.06 |
| Queries that were completely incorrect | 25 (30.1) | 15 (18.1) | 0.07 |
| Answers with medications not on the list provided | 2 (2.4) | 0 | 0.16 |
| Queries where percentage of correct answers decreased with web-assisted query compared to internal query | | 24 (28.9) | |
| Queries where percentage of correct answers increased with web-assisted query compared to internal query | | 35 (42.2) | |
| Category C interactions (n=30) | | | |
| Percentage of correct answers | 0 (0-55) | 100 (60-100) | <0.01 |
| Queries with >1 unique answers | 9 (30) | 7 (23.3) | 0.56 |
| Queries with >2 unique answers | 4 (13.3) | 1 (3.3) | 0.16 |
| Queries that were completely correct | 7 (23.3) | 19 (63.3) | <0.01 |
| Queries that were completely incorrect | 20 (66.7) | 5 (16.7) | <0.01 |
| Category D interactions (n=23) | | | |
| Percentage of correct answers | 80 (40-100) | 100 (100-100) | 0.01 |
| Queries with >1 unique answers | 13 (56.5) | 2 (8.7) | <0.01 |
| Queries with >2 unique answers | 4 (17.4) | 0 | 0.04 |
| Queries that were completely correct | 10 (43.5) | 20 (87) | <0.01 |
| Queries that were completely incorrect | 2 (8.7) | 1 (4.3) | 0.55 |
| Category X interactions (n=6) | | | |
| Percentage of correct answers, | 40 (5-75) | 100 (85-100) | 0.08 |
| Queries with >1 unique answers | 3 (50) | 1 (16.7) | 0.22 |
| Queries with >2 unique answers | 2 (33.3) | 0 | 0.14 |
| Queries that were completely correct | 1 (16.7) | 4 (66.7) | 0.08 |

| Queries that were completely incorrect | 2 (33.3) | 0 | 0.14 |
|---|---|---|---|
| No interactions (n=24) | | | |
| Percentage of correct answers | 100 (70-100) | 40 (0-60) | <0.01 |
| Queries with >1 unique answers | 8 (33.3) | 17 (70.8) | <0.01 |
| Queries with >2 unique answers | 1 (4.2) | 4 (16.7) | 0.16 |
| Queries that were completely correct | 15 (62.5) | 2 (8.3) | <0.01 |
| Queries that were completely incorrect | 1 (4.2) | 9 (37.5) | <0.01 |

IQR: interquartile range, SD: standard deviation

*used gpt-4o model

**used gpt-4o-search-preview

Data reported as n (%) or median (IQR) unless otherwise specified

Each query (using internal knowledge and web-assisted) was asked 5 times- “unique” answers are each pair of medications that was selected as the drug-drug interaction pair or “no interaction”. If the drug-drug interaction pair was “meropenem and valproic acid” and the internal-knowledge query answered “meropenem and valproic acid” four times but “no interaction” one time, this would be considered to have 2 unique answers and to be partially correct, as it did not select the correct answer 100% of the time (completely correct) or 0% of the time (completely incorrect).